\RequirePackage[OT1]{fontenc}

\RequirePackage{mathtools}
\RequirePackage{siunitx}
\RequirePackage[pdfpagelabels=false, hidelinks]{hyperref}
\RequirePackage[nameinlink]{cleveref}
\RequirePackage{xspace}

\Crefname{equation}{Eq.}{Eqs.}
\Crefname{figure}{Fig.}{Figs.}

\documentclass[letterpaper, 10 pt, conference]{ieeeconf}  

\IEEEoverridecommandlockouts                              

\title{\LARGE \bf
Current-Based Impedance Control for Interacting\\ with Mobile Manipulators
}

\author{Jelmer de Wolde$^{1}$, Luzia Knoedler$^{1}$, Gianluca Garofalo$^{2}$, and Javier Alonso-Mora$^{1}$
\thanks{$^{1}$ Cognitive Robotics (CoR) department,
        Delft University of Technology, 2628 CD Delft, The Netherlands
    {\tt\small j.s.dewolde@student.tudelft.nl, \{l.knoedler, j.alonsomora\}@tudelft.nl}}%
\thanks{$^{2}$ ABB Corporate Research, 72178 V{\"a}ster{\aa}s, Sweden
        {\tt\small gianluca.garofalo@se.abb.com}}%
\thanks{This paper has received funding from the European Union's Horizon 2020 research and innovation programme under grant agreement No. 101017008. All
content represents the opinion of the author(s), which is not necessarily shared or endorsed by their respective employers and/or sponsors.}}

\usepackage{bm}
\usepackage{derivative}
\usepackage{subcaption}
\usepackage{stfloats} 
\usepackage{amssymb}
\usepackage{url}
\usepackage[acronym]{glossaries}
\newacronym{com}{COM}{Center of Mass}

\newacronym{lw}{lw}{lower wrist}
\newacronym{uw}{uw}{upper wrist}
\newacronym{fa}{fa}{forearm}
\newacronym{ua}{ua}{upper arm}

\newacronym{ee}{EE}{End-Effector}
\usepackage{pgfplots}
\usepackage{tikzscale}
\usetikzlibrary{external, decorations.markings, arrows.meta, pgfplots.statistics}
\pgfplotsset{compat=1.18}

\DeclareMathOperator{\sign}{sign}

\newcommand{\q}{\ensuremath{\bm{q}}\xspace}
\newcommand{\dq}{\ensuremath{\dot{\bm{q}}}\xspace}
\newcommand{\ddq}{\ensuremath{\ddot{\bm{q}}}\xspace}

\newcommand{\dx}{\ensuremath{\dot{\bm{x}}}\xspace}
\newcommand{\ddx}{\ensuremath{\ddot{\bm{x}}}\xspace}

\newcommand{\e}{\ensuremath{\bm{e}}\xspace}
\newcommand{\de}{\ensuremath{\dot{\bm{e}}}\xspace}

\newcommand{\M}{\ensuremath{\bm{M}(\q)}\xspace}
\newcommand{\C}{\ensuremath{\bm{C}(\q, \dq)}\xspace}
\newcommand{\g}{\ensuremath{\bm{\tau}_\mathrm{g}(\q)}\xspace}
\newcommand{\f}{\ensuremath{\bm{\tau}_\mathrm{f}}\xspace}

\newcommand{\Mo}{\ensuremath{\bm{\Lambda}(\q)}\xspace}
\newcommand{\Co}{\ensuremath{\bm{\mu}(\q, \dq)}\xspace}
\newcommand{\go}{\ensuremath{\bm{f}_\mathrm{g}(\q)}\xspace}
\newcommand{\fo}{\ensuremath{\bm{f}_\mathrm{f}}\xspace}

\newcommand{\J}{\ensuremath{\bm{J}(\q)}\xspace}
\newcommand{\JT}{\ensuremath{\bm{J}^T(\q)}\xspace}

\newcommand\Remember[2]
{
    \expandafter\gdef\csname labeled:#1\endcsname{#2}#2
}
\newcommand\Recall[1]
{
    \csname labeled:#1\endcsname
}
\IfFileExists{local_pc}{\tikzexternalize[prefix=tikz_figures/]}{\tikzexternalize[]}
\begin{document}

\maketitle
\thispagestyle{empty}
\pagestyle{empty}

\begin{abstract}

    As robots shift from industrial to human-centered spaces, adopting mobile manipulators, which expand workspace capabilities, becomes crucial. In these settings, seamless interaction with humans necessitates compliant control. Two common methods for safe interaction, admittance, and impedance control, require force or torque sensors, often absent in lower-cost or lightweight robots. This paper
    presents an adaption of impedance control that can be used on current-controlled robots without the
    use of force or torque sensors and its application for compliant control of a mobile manipulator. A calibration method is designed that enables estimation of the
    actuators' current/torque ratios and frictions, used by the adapted impedance controller, and that can handle model errors. The
    calibration method and the performance of the designed controller are experimentally validated
    using the Kinova GEN3 Lite arm. Results show that the calibration method is consistent and that the
    designed controller for the arm is compliant while also being able to track targets with five-millimeter
    precision when no interaction is present.
    Additionally, this paper presents two operational modes for interacting with the mobile manipulator: one for guiding the robot around the workspace through interacting with the arm and another for executing a tracking task, both maintaining compliance to external forces. These operational modes were tested in real-world experiments, affirming their practical applicability and effectiveness.
    
    \vspace{0.1cm}
    \footnotesize{\textbf{Code: \urlstyle{same}\url{https://github.com/tud-amr/mobile-manipulator-compliance}}}
\end{abstract}
\vspace{0.5cm}

\section{INTRODUCTION}
As robots transition from industrial settings to human-centered spaces, integrating mobile manipulation and compliant control becomes vital for efficient and safe task execution~\cite{zacharaki2020safety}. 
Mobile manipulation extends the robot's workspace by combining navigation and manipulation capabilities, enhancing flexibility and efficiency.
Moreover, compliance is mandatory for robust behaviors in unstructured environments and safe operation in the proximity of humans~\cite{siciliano2008springer}. Compliance is commonly achieved through indirect force control methods such as admittance and impedance control. 
\begin{figure}
    \centering
\includegraphics[width=1\columnwidth]{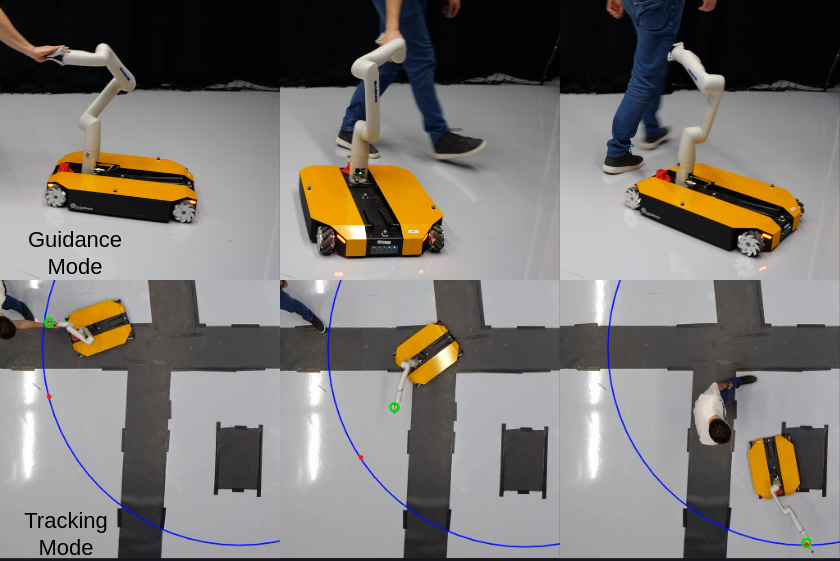}
    \caption{Compliance-enabled operational modes implemented on mobile manipulator without force/torque sensors.
    Guidance mode: mobile manipulator is led through interaction with arm.
    Tracking mode: end-effector~(green circle) tracks target~(red dot) while being compliant with user interactions.}
    \label{fig:results_exp3}
\end{figure}
However, these methods require contact information that are typically derived from measurements of the joint torques or from a force/torque sensor mounted at the end-effector. Such equipment can be expensive, and off-the-shelf robots might not come equipped with them. 
Further challenges arise based on the control mode supported by the robot.
On the one hand, while admittance control can be applied to position/velocity-controlled robots, the robot is only compliant with measured forces.
On the other hand, impedance control effectively achieves compliance to all forces interacting with the robot but is exclusive to torque-controlled robots.

Depending on the capabilities of the manipulator and the mobile base, either whole-body impedance control~\cite{bussmannWholeBodyImpedanceControl2018-05}, admittance control for the base, and impedance control for the arm~\cite{dietrich2016whole, kimMOCAMANMObileReconfigurable2020-05, kimWholebodyControlNonholonomic2019-07, lamonIntelligentCollaborativeRobotic2020-05}, or whole-body admittance control~\cite{leboutetTactilebasedComplianceHierarchical2016-11} have been applied.
Whole-body impedance control is less often employed since off-the-shelf mobile bases typically do not provide torque control. Furthermore, the base's weight and motor friction can hinder interaction with the base.
To implement impedance control on the arm, torque sensors at each joint are necessary. However, lightweight or more affordable robots often lack torque sensors and may only support current control.
Admittance control faces the challenge of needing sensors to detect external forces on each link that require compliant behavior.

Hence, implementing compliant control on a mobile manipulator poses significant challenges, particularly when using off-the-shelf manipulators and wheeled mobile bases, as some do not provide the required sensors or control modes. 
In the remainder of the paper, we will generically use the term force/torque sensor to refer to such sensing capability.

Thus, this work presents a holistic approach to address compliance for mobile manipulators using off-the-shelf current-controlled manipulators and velocity-controlled mobile bases without adding force/torque sensors.
In the context of the manipulator, 
the approach involves estimating the direct correlation between
the actuator current and the resulting joint torque, overcoming the typical reliance of impedance control on torque sensors. 
Additionally, we present two operational modes for interacting with the mobile manipulator see \Cref{fig:results_exp3}. A \textit{guidance mode} to guide the mobile manipulator around the workspace through interacting with the arm, and a \textit{tracking mode} to execute a tracking task. Both modes are compliant in the case of contacts.

In summary, our contributions are:
\begin{itemize}
    \item A calibration method that estimates the actuators' current/torque ratios and friction of the manipulator's joints while handling model inaccuracies, enabling current-based impedance control without requiring any additional sensing beyond standard robot control systems.
    \item Two different operational modes for interacting with mobile manipulators through the compliant arm.
    \item Real-world experimental validation of the calibration method, the performance of the adapted impedance controller for the manipulator, and the mobile manipulation operational modes.
\end{itemize}

\section{RELATED WORK}
We start with existing implementations of compliant control on mobile manipulators, highlighting their connection with the used hardware. 
Additionally, we explore how past approaches have tackled hardware constraints.

Whole-body impedance control was implemented on the DLR Lightweight Rover Unit as it supports torque control for both the arm and the base while considering the nonholonomic constraints of the base~\cite{bussmannWholeBodyImpedanceControl2018-05}. The resulting active compliant base behavior is particularly beneficial as it allows compliance to unmeasured forces acting on the base.  This is only possible for torque-controlled back-drivable wheels.

In contrast, off-the-shelf mobile bases are usually position/velocity-controlled. 
Thus, \cite{dietrich2016whole} implements an admittance interface on the Rollin' Justin to transform the desired forces and torques into applicable motion trajectories. The authors provide a formal stability proof, which is necessary when combining torque control and admittance control.

Finally, whole-body admittance control can be applied to completely position/velocity-controlled mobile manipulators. This requires the measurement of the external forces. Therefore, the research for these platforms is mainly focused on alternatives to force/torque sensors such as moment-observers~\cite{xing2021admittance} and to enable sensing for other parts of the robot than the \gls{ee}~\cite{leboutetTactilebasedComplianceHierarchical2016-11}. 

In general, a substantial body of research has focused on the estimation of external forces and torques without the use of sensors. These are almost exclusively model-based and thus are highly dependent on the accuracy of the dynamics model. The external wrench is considered as a disturbance and estimated by comparing the nominal and the observed model output \cite{eom1998disturbance, alcocer2003force, jung2006robust}.
Some works focus on the case of current-controlled robots using Kalman filter \cite{wahrburg2017motor} or learning-based methods~\cite{liu2021sensorless}.
In~\cite{wahrburg2017motor}, gearbox ratio and motor constant are assumed known.
Moreover,~\cite{wahrburg2018modeling} provide an identification procedure for different friction effects for motor-controlled robots.
In \cite{cao2022contact} a simple calibration method is applied to estimate the current/torque and gearbox ratio, but assume an accurate dynamic model.

In this work, we offer a comprehensive description of the calibration method leveraging the known gravitational force on the robot to estimate the current/torque ratio and friction loss. Additionally, we enhance the method to address model errors. This allows the application of current-based impedance control on a current-controlled manipulator. Furthermore, we combine the compliant manipulator with a mobile base and develop two operational modes for interacting with the mobile manipulator through the arm.

\section{PRELIMINARIES}
This section offers a concise introduction to the robot dynamics and impedance control and formulates the problem considered in this work.

\subsection{Robot Dynamics}
The dynamics of a manipulator with $n$ rigid joints  are represented by the nonlinear differential equation:
\begin{equation}
    \label{eq:dynamics_q}
    \M \ddq + \C \dq + \g + \f = \bm{\tau} + \bm{\tau}_\mathrm{ext},
\end{equation}
where the symmetric, positive definite inertia matrix $\M \in \mathbb{R}^{n \times n}$, the Coriolis/centrifugal matrix $\C \in \mathbb{R}^{n \times n}$, and the gravitational torque $\g \in \mathbb{R}^n$ depend on the joint positions $\q \in \mathbb{R}^n$ and the joint velocities $\dq \in \mathbb{R}^n$. The torques applied to the system are the actuator torques $\bm{\tau} \in \mathbb{R}^n$ and the external torques $\bm{\tau}_\mathrm{ext} \in \mathbb{R}^n$. The vector $\f \in \mathbb{R}^n$ contains the torques due to friction.

Given the task coordinates $\bm{x}$ and $\dx$, each term in \eqref{eq:dynamics_q} can be transformed in task space \cite{ott2008cartesian}, leading to the model:
\begin{equation}
    \Mo \ddx + \Co \dx + \go + \fo = \bm{f} + \bm{f}_\mathrm{ext},
\end{equation}
with $\Mo$, $\Co$ being the Cartesian inertia matrix and Coriolis/centrifugal matrix, respectively. The effects of gravity, friction, actuator, and external forces are given by $\go$, $\fo$, $\bm{f}$, and $\bm{f}_\mathrm{ext}$, respectively.

\subsection{Impedance Control}
Impedance control aims to let the \gls{ee} of the robot behave as a spring-damper system with the desired stiffness~$\bm{K}_d$ and damping~$\bm{D}_d$. In operational space, an impedance controller~\cite{ott2008cartesian}~\cite{dietrich2021practical} is generally defined as:
\begin{equation}
    \label{eq:impedance_controller}
    \bm{f} = \underbrace{\Mo \ddx_\mathrm{d} + \Co \dx_\mathrm{d} + \bm{f}_\mathrm{g}}_\text{"+"-part} \underbrace{- \bm{K}_\mathrm{d} \e - \bm{D}_\mathrm{d} \de \vphantom{\Mo \ddx_\mathrm{d} + \Co \dx_\mathrm{d} + \bm{f}_\mathrm{g}}}_\text{PD-part},
\end{equation}
where the "+"-part compensates for the effect of gravity and adds feedforward terms. The PD component specifies the desired compliance of the robot concerning the position and velocity errors of the \gls{ee}, represented as $\bm{e} = \bm{x} - \bm{x}_\mathrm{d}$ and $\dot{\bm{e}}$, respectively, where $\bm{x}_\mathrm{d}$ denotes the desired \gls{ee} position.
The required joint torques can be computed using the Jacobian matrix $\J$, with ${ \dx = \J \dq }$ and:
\begin{equation}
    \label{eq:impedance_torque}
    \bm{\tau} = \JT\bm{f}.
\end{equation}
In case of redundancy, the general concept of nullspace projection \cite{dietrich2015overview} can be used to add secondary task joint torques $\bm{\tau}_2$, projected using the projection matrix $\bm{N}$, i.e.:
\begin{equation}
    \label{eq:impedance_torque_ns_task}
    \bm{\tau} = \JT\bm{f} + \bm{N} \bm{\tau}_2.
\end{equation}

\subsection{Problem Formulation}
We consider a mobile manipulator, including a current-controlled manipulator without force/torque sensors, which does not allow the direct application of \Cref{eq:impedance_torque_ns_task}. Thus, the goal is to derive the relation
\begin{equation}
    \label{eq:ct_relation}
   \bm{c}= \bm{h}(\bm{\tau})
\end{equation}
that matches the actuator currents $\bm{c}$ with the joint torques from the impedance controller $\bm{\tau}$.
Additionally, the objective includes integrating this manipulator arm with a mobile base by defining a control method for the base.
\begin{figure}[!b]
    \centering
    \begin{minipage}{.5\columnwidth}
        \centering
        \includegraphics[height=1.2\columnwidth]{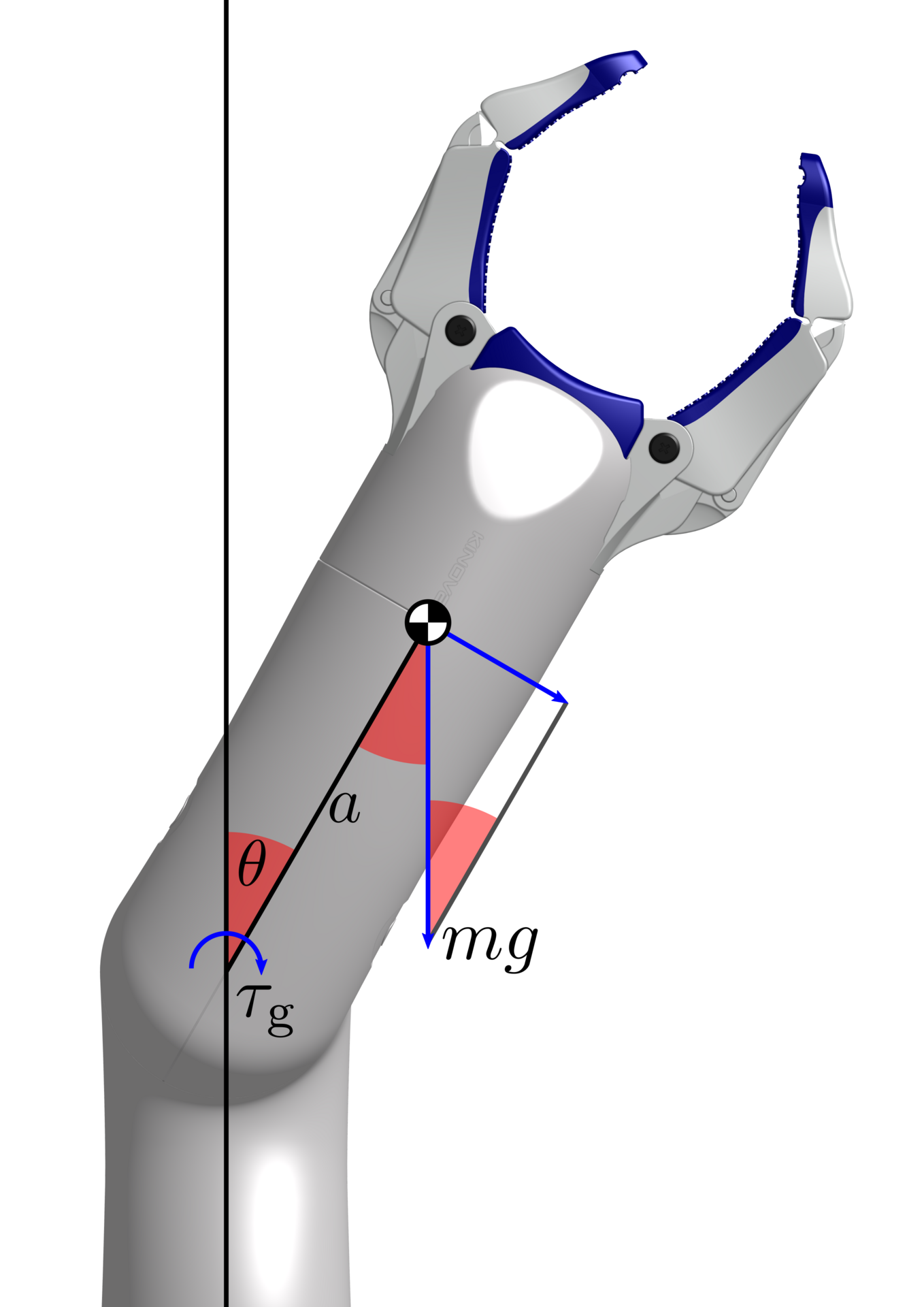}
        \caption{Torque acting on a \\ joint due to gravity.}
        \label{fig:joint_correct_com}
    \end{minipage}%
    \begin{minipage}{.5\columnwidth}
        \centering
        \includegraphics[height=1.2\columnwidth]{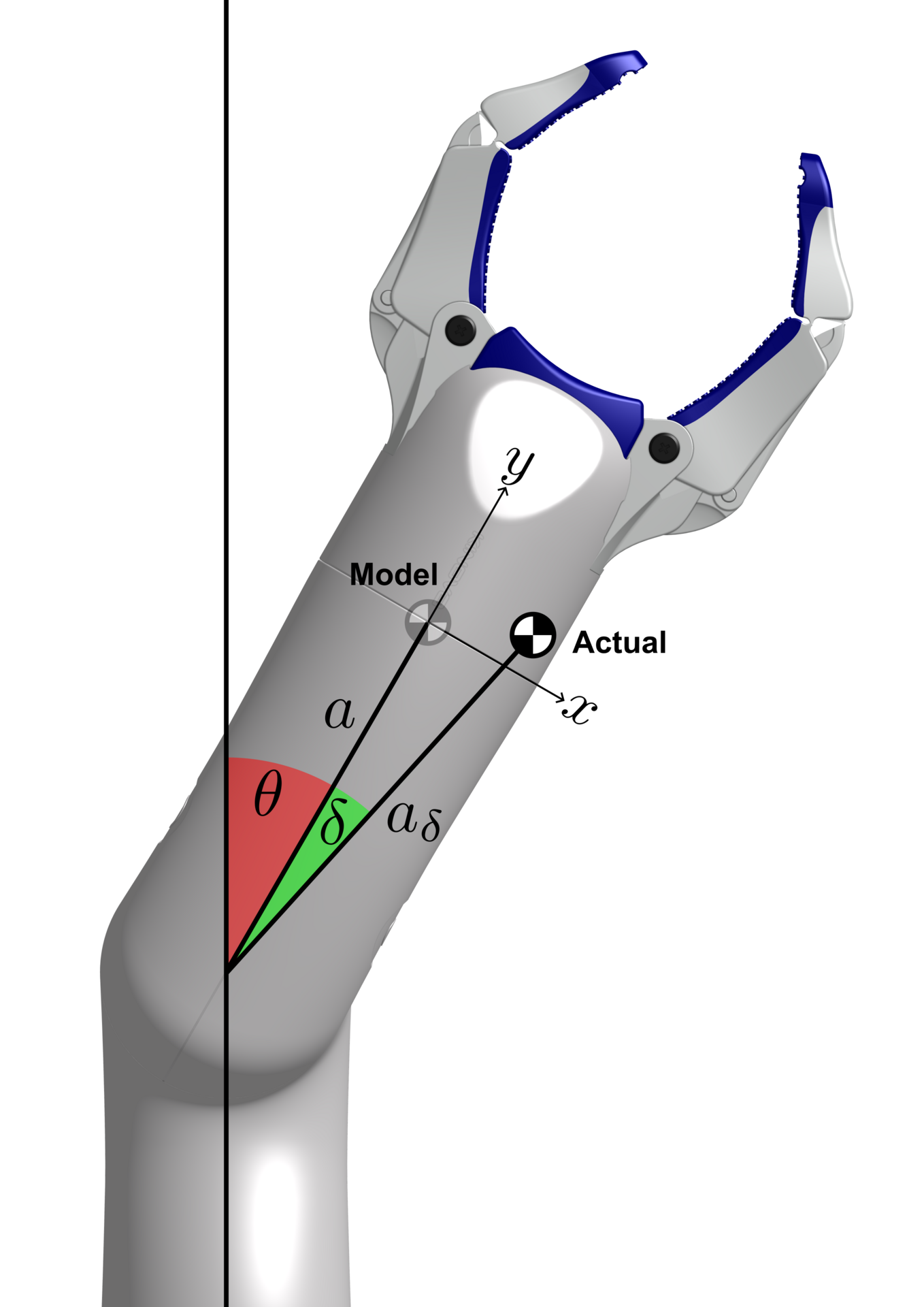}
        \caption[Difference between modelled and actual \acrshort{com}.]
        {Difference between modelled and actual \acrshort{com}.}
        \label{fig:inaccuracy_com}
    \end{minipage}
\end{figure}

\section{METHOD}
\label{section:method}
This section describes our main contribution: a calibration method to implement impedance control on a lightweight off-the-shelf manipulator without force/torque sensors in \Cref{subsection:calibration}, and two operational modes for interacting with the mobile manipulator in \Cref{subsection:base}.

\subsection{Calibration}
\label{subsection:calibration}
We develop a calibration process to establish the relation in \Cref{eq:ct_relation}, detailed in \Cref{subsection:ratio}, to implement current-based impedance control without torque sensors. This process utilizes the robot's model and Earth's gravity to estimate the ratio between the actuator current and the resulting torque. Additionally, it estimates joint friction (explained in \Cref{subsection:friction}) and incorporates a compensatory control strategy. Challenges arising from model inaccuracies and their resolution are addressed in \Cref{subsection:phase_shift}. %
To enhance clarity, we outline the process for a single joint below but note that this can be replicated for each joint.

\subsubsection{Current/Torque Ratio Estimation}
\label{subsection:ratio}
\begin{figure}[!b]
    \centering
    \includegraphics[width=\columnwidth, height=0.6\columnwidth]{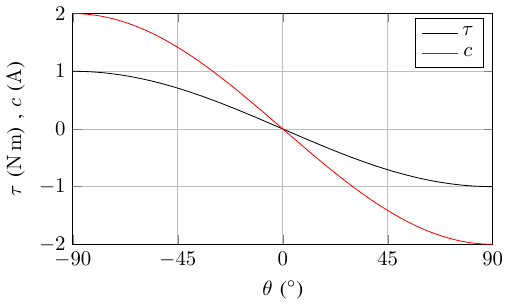}
    \caption{Torque and current for a joint with $K=1$ and a current/torque ratio of $r = 2 \si{\ampere\per\newton\per\meter}$.}
    \label{fig:calibration_ratio}
\end{figure}

We start by estimating the ratio $r$ between the actuator current $c$ and the model-based actuator torque $\tau$ so that the relation from \Cref{eq:ct_relation} for a single joint becomes 
\begin{equation}
    \label{eq:controller_ratio}
    c = r \tau.
\end{equation}
Initially, we assume zero friction, but friction is addressed in \Cref{subsection:friction}. 
In an ideal case, the recorded current for a joint with $r= 2$ assuming zero friction would mimic the red plot illustrated in Figure \ref{fig:calibration_ratio}. 
For any given robot pose, it is possible to calculate the gravitational torque $\tau_\mathrm{g}$ acting on the joint. The dynamic equation of the rotation of a single joint can be expressed as:
\begin{equation} \label{eq:dynamic_eqation_rotation}
    I \alpha = \tau_\mathrm{g} + \tau,
\end{equation}

\noindent
where $I$ is the inertia of the body rotating around the joint, $\alpha$ is the angular
acceleration, and $\tau$ is the actuator torque. Note, that all other joints are locked, and we consider single-axis joints, resulting in $I$ being a constant value for the considered joint. If the robot's body rotates around the joint with a constant velocity, so
that $\alpha = 0$, \Cref{eq:dynamic_eqation_rotation} can be rewritten as:
\begin{equation} \label{eq:torques_no_friction}
    \tau = -\tau_\mathrm{g}.
\end{equation}

\noindent
Therefore, for a constant rotational velocity of the robot, $\tau$
is exactly the opposite of $\tau_\mathrm{g}$, which depends on the mass $m$ the joint is carrying, the gravity constant $g$, the length of the moment arm $a$, and the angle between the moment arm and the direction of gravity $\theta$, as
shown in \Cref{fig:joint_correct_com}. The gravitational torque on a joint is therefore defined
as:
\begin{equation} \label{eq:gravitational_torque}
    \tau_\mathrm{g} = m g a \sin(\theta).
\end{equation}

\noindent
In this equation, $m g a$ is a constant factor. For simplicity, we set $m g a$ to
$K$ during the explanation of the current/torque ratio and friction estimation. This results in  $\tau_\mathrm{g} = K\sin(\theta)$. Thus, when rotating the joint from $\theta = -90 \si{\degree}$
to $\theta = 90 \si{\degree}$ with a constant velocity, $\tau = -K\sin(\theta)$ holds as plotted in
black in \Cref{fig:calibration_ratio} for $K=1$.

We use the default, non-compliant controller of the robot to rotate the joint from $\theta = -90 \si{\degree}$
to $\theta = 90 \si{\degree}$ with a constant velocity and record the applied actuator current~$c$. This results in the dataset ~$\mathcal{D}$ containing~$N$ measurements of $c$ and the corresponding model-based actuator torque~$\tau$ computed using \Cref{eq:torques_no_friction}:
\begin{equation}
    \mathcal{D} = \left\{ \left(c_1, \tau_1 \right), \ldots, \left(c_N, \tau_N \right) \right\}.
\end{equation}
Using $\mathcal{D}$, it is possible to estimate the ratio $r$. This can be done by solving the following optimization problem:
\begin{equation} \label{eq:minimization_ratio}
    r^* = \operatorname*{argmin}_{r} \sum_{i=1}^{N} (c_i - r \tau_i)^2,
\end{equation}
\noindent
where $c_i$ and $\tau_i$ are the measured current and the model-based torque of sample $i$, respectively.

\begin{figure}[!b]
    \centering
    \includegraphics[width=\columnwidth, height=0.6\columnwidth]{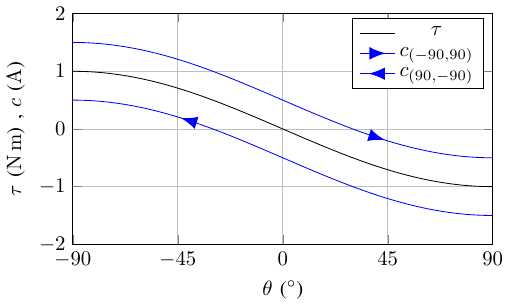}
    \caption{Torque and current for a joint with $K=1$, $r=1$ and a friction loss of $l = 0.5 \si{\ampere}$.}
    \label{fig:calibration_friction}
\end{figure}

\subsubsection{Friction Loss Estimation}
\label{subsection:friction}

The calibration in \Cref{subsection:ratio} assumes a frictionless system, yet real-world scenarios inevitably involve friction, which works against the movement of the joint. To simplify and isolate the effect of friction, we consider an actuator with $r=1$ during the explanation.
\Cref{fig:calibration_friction} shows again an idealized scenario with $\tau$ in black and $c$ in
blue when $0.5 \si{\ampere}$ of the current is constantly lost due to
friction. In contrast to \Cref{fig:calibration_ratio}, \Cref{fig:calibration_friction} shows two lines
for $\tau$, depending on the direction of the movement. The actuator current is shifted up relative to $\tau$ for a movement from $\theta = -90 \si{\degree}$ to $\theta = 90 \si{\degree}$, while it is shifted down for the opposite movement from $\theta = 90
\si{\degree}$ to $\theta = -90 \si{\degree}$. 
This occurs because friction opposes upward link movement initially, causing the current to exceed the model torque. Conversely, during downward movement, friction hinders joint acceleration, leading to the current being lower than the model torque.
Using $\mathcal{D}$, it is possible to estimate $r$ and $l$, by solving the following minimization problem:
\begin{equation}\label{eq:minimization_ratio_and_friction}
    r^*, l^* = \operatorname*{argmin}_{r, l} \sum_{i=1}^{N} (c_i - (r \tau_i + l))^2,
\end{equation}

\noindent
where $c_i$ and $\tau_i$ are the measured current and model-based torque of sample $i$, respectively. The minimization problem can be solved with a solver supporting multiple variables, for which we employ Python's Scipy library. Depending on the direction of movement during the collection of the data points, the resulting friction loss $l$ can be positive or negative, but the absolute value is of interest. 

\subsubsection{Friction Loss Compensation}
\label{subsection:friction_compensation}

To enable the controller to compensate for the estimated friction loss $l$, we assume the value obtained through calibration remains constant. We can add friction compensation to \Cref{eq:controller_ratio} by adding the friction loss value in the direction of the original torque output:
\begin{equation} \label{eq:controller_fc_torque}
    c = r \tau + \textup{sign}(\tau) l.
\end{equation}

\noindent
However, compensating for the friction using this method on its own results in a less compliant robot.
Pushing away the robot will now require a force large enough to overcome the friction twice, once
for the actual friction and once since the friction loss compensation of the controller is also always
pushing the robot back to its target. Another option is to compensate for friction loss in the direction
of moving, which makes sense since friction is always directed in the opposite direction of the
velocity:
\begin{equation} \label{eq:controller_fc_velocity}
    c = r \tau + \textup{sign}(\dot{q}) l.
\end{equation}

\noindent
Nevertheless, this approach fails when the robot remains at zero velocity, and the position error is insufficient to overcome friction and initiate movement. A solution is to use a
combination of both methods, where the contribution of each depends on the velocity of the joint,
where only the velocity-based compensation is used for an absolute joint velocity above a threshold
velocity $t$, while only torque-based compensation is used for zero velocity:
\begin{equation} \label{eq:controller_fc}
    c = r \tau + l \left(\min\left[\frac{|\dot{q}|}{t}, 1\right] \left(\sign \dot{q} - \sign \tau \right) + \sign \tau \right).
\end{equation}

\begin{figure}[!b]
    \centering
    \includegraphics[width=\columnwidth, height=0.6\columnwidth]{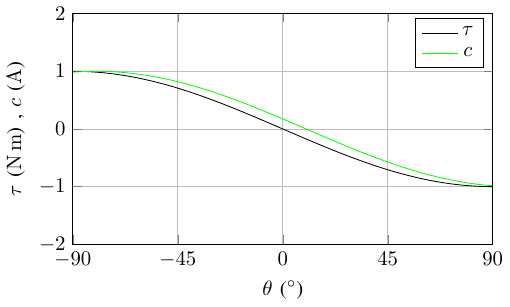}
    \caption{Torque and current for a joint with $K=1$, $r=1$, $l=0$ and a phase shift of $\delta = -10 \si{\degree}$.}
    \label{fig:calibration_phase}
\end{figure}

\subsubsection{Phase Shift Estimation}
\label{subsection:phase_shift}

The approach described for deriving the ratio $r$ and friction loss $l$ relies entirely on the accuracy of the robot's model.  However, inaccuracies in the model, as depicted in \Cref{fig:inaccuracy_com}, can occur, where the \gls{com} is displaced. A displacement in the $z$ direction would
mean a displacement parallel to the axis of rotation of the joint and therefore does not affect the
calibration of this joint. A displacement in the $y$ direction would affect the length of the
moment arm $a$, but since this is part of the constant factor $mga$, only the current/torque ratio
will be slightly affected. A displacement in the $x$ direction leads to major problems since this
results in a reformulation of the gravitational torque from \Cref{eq:gravitational_torque} to:
\begin{equation}
    \tau_\mathrm{g} = m g a_{\delta} \sin(\theta + \delta).
\end{equation}

\noindent
Here, $a_{\delta}$ represents the actual moment arm of the robot, which is different from the moment arm of the model $a$. The discrepancy in angle between the model and the actual robot is denoted by $\delta$. To focus only on this effect, let us assume to have an idealized actuator with $r=1$ and $l=0$. We would then obtain a phase difference between the
model-based actuator torque $\tau$ and the measured current $c$ as in \Cref{fig:calibration_phase}, where $\tau$ is shown in black and $c$ in green for an angle difference of $\delta = -10 \si{\degree}$.

\begin{figure}[!b]
    \centering
    \includegraphics[height=0.5\columnwidth]{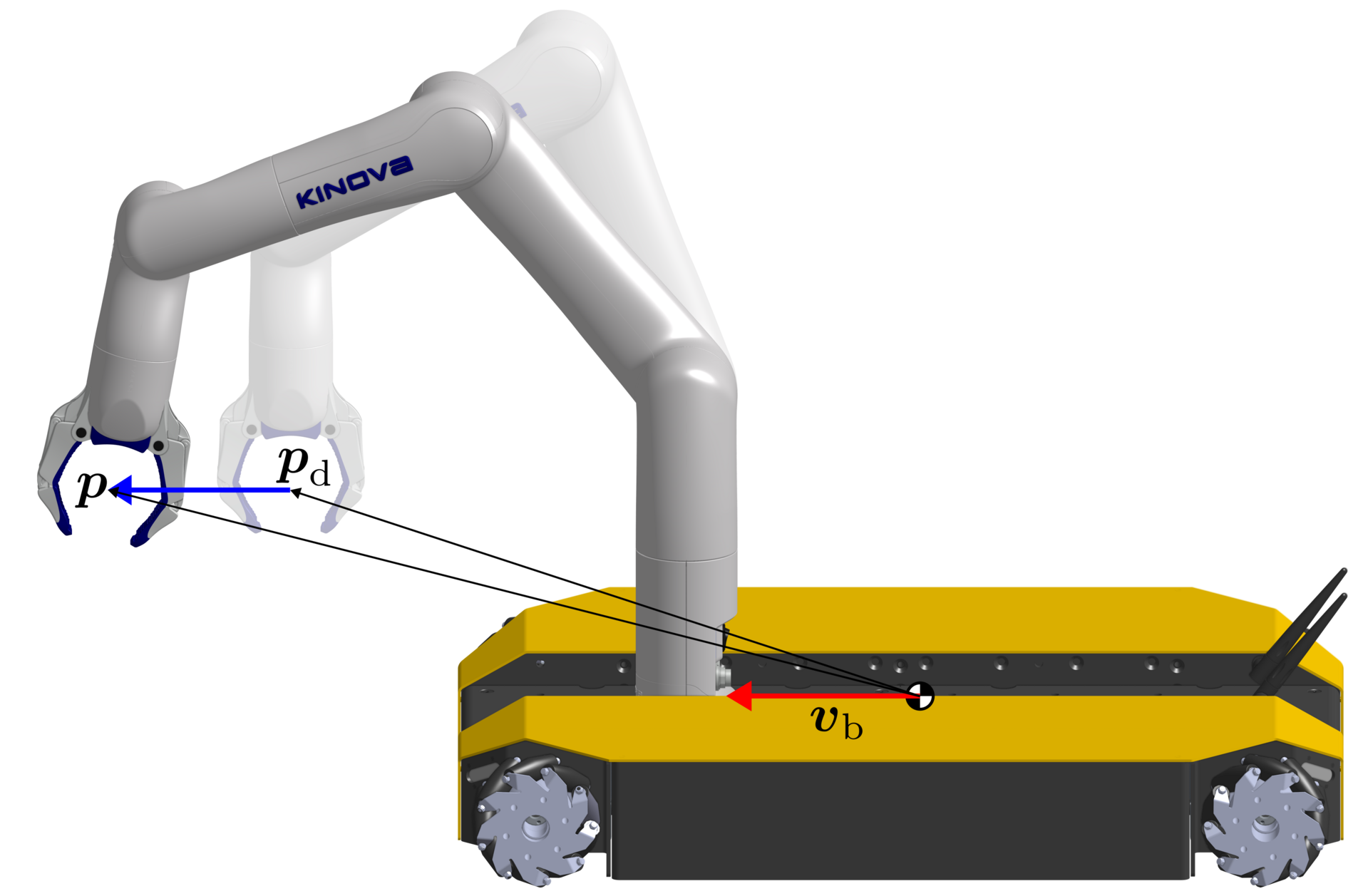}
    \caption[Functioning of the base controller.]
    {The base follows the end-effector.}
    \label{fig:base_control}
\end{figure}

If there exists a phase difference, the estimation of $r$ and $l$ as
described in \Cref{eq:minimization_ratio_and_friction} will be inaccurate. To solve this problem,
the robot model must be adjusted to match the real robot. Considering that the calculation of gravitational torque on a joint relies on the combined \gls{com} locations of all individual links, adjustments should be made starting from the last link and proceeding backward to the first. 
To estimate the angle difference $\delta$, we record the joint angle $\theta$ next to $c$ resulting in the new dataset:
\begin{equation}
    \mathcal{D} = \left\{ \left(c_1, \theta_1 \right), \ldots, \left(c_N, \theta_N \right) \right\}.
\end{equation}

\noindent
Note that if the joint is also influenced by friction, a vertical displacement of the current data
is caused, as was shown in \Cref{fig:calibration_friction}. By recording the data on the trajectory
from ${\theta = -90 \si{\degree}}$ to ${\theta = 90 \si{\degree}}$ and on the trajectory from ${\theta = 90 \si{\degree}}$ to ${\theta = -90 \si{\degree}}$ and taking the average on each angle $\theta$, the
effect of friction will be canceled out. Next, it is possible to solve the following minimization
problem:
\begin{equation} \label{eq:gravitational_torque_shift}
    s^*, \delta^* = \operatorname*{argmin}_{s, \delta} \sum_{i=1}^{N} (c_i - s \sin(\theta_i + \delta))^2,
\end{equation}

\noindent
where $c_i$ and $\theta_i$ are the measured actuator current and the joint
rotation at sample $i$, respectively, and $s$  is a scaling factor combining $mgar$.
For $\delta\not\approx 0$, the model of the robot should be adapted by moving the \gls{com} location to align with the estimated angle difference $\delta$. Note that the scaling factor
$s$ is not used as the original minimization problem from \Cref{eq:minimization_ratio_and_friction} can be used after correcting the model.

\subsection{Operational Modes for a Mobile Manipulator}
\label{subsection:base}
We have introduced a calibration method that allows impedance control on a current-controlled manipulator without force/torque sensors.
Implementing impedance control on many off-the-shelf mobile bases is not possible since they only allow position/velocity control.
This is also the case for the considered mobile base. If current control was available, this would require a similar calibration method as presented in \Cref{subsection:calibration}. Another challenge is that the base is difficult to push into a desired direction due to its omnidirectional wheels, even with the motors turned off. 

Hence, an alternative control strategy is employed for the base, where the base controller aims to keep the \gls{ee} position $\bm{p}$ always at a defined desired position $\bm{p}_\mathrm{d}$, both expressed in the reference frame of the base itself as shown in \Cref{fig:base_control}. A difference between $\bm{p}$ and $\bm{p}_\mathrm{d}$ results in a base velocity $\bm{v}_\mathrm{b}$ using a defined constant gain $\bm{K}_\mathrm{b}$:
\begin{equation} \label{eq:base_controller}
    \Remember{eq: base-controller}
    {
        \bm{v}_\mathrm{b} = \bm{K}_\mathrm{b} (\bm{p} - \bm{p}_\mathrm{d}).
    }
\end{equation}

\noindent
With this controller, the base will always follow the \gls{ee} when the \gls{ee} deviates from the desired position $\bm{p}_\mathrm{d}$. Note that a deviation in the vertical direction has no effect, as it's evident that the base can only move within the horizontal plane of the ground. Using the omnidirectional relation between wheel motions and the resulting motion of the base, the correct wheel velocities can be produced that correspond with the direction and magnitude of the desired base velocity $\bm{v}_\mathrm{b}$.

\begin{figure}[!b]
    \centering
    \def\datafile{data/calibration/calibrate_torque_and_friction.csv}
    \begin{subfigure}{.5\columnwidth}
        \centering
        \includegraphics[width=0.95\textwidth]{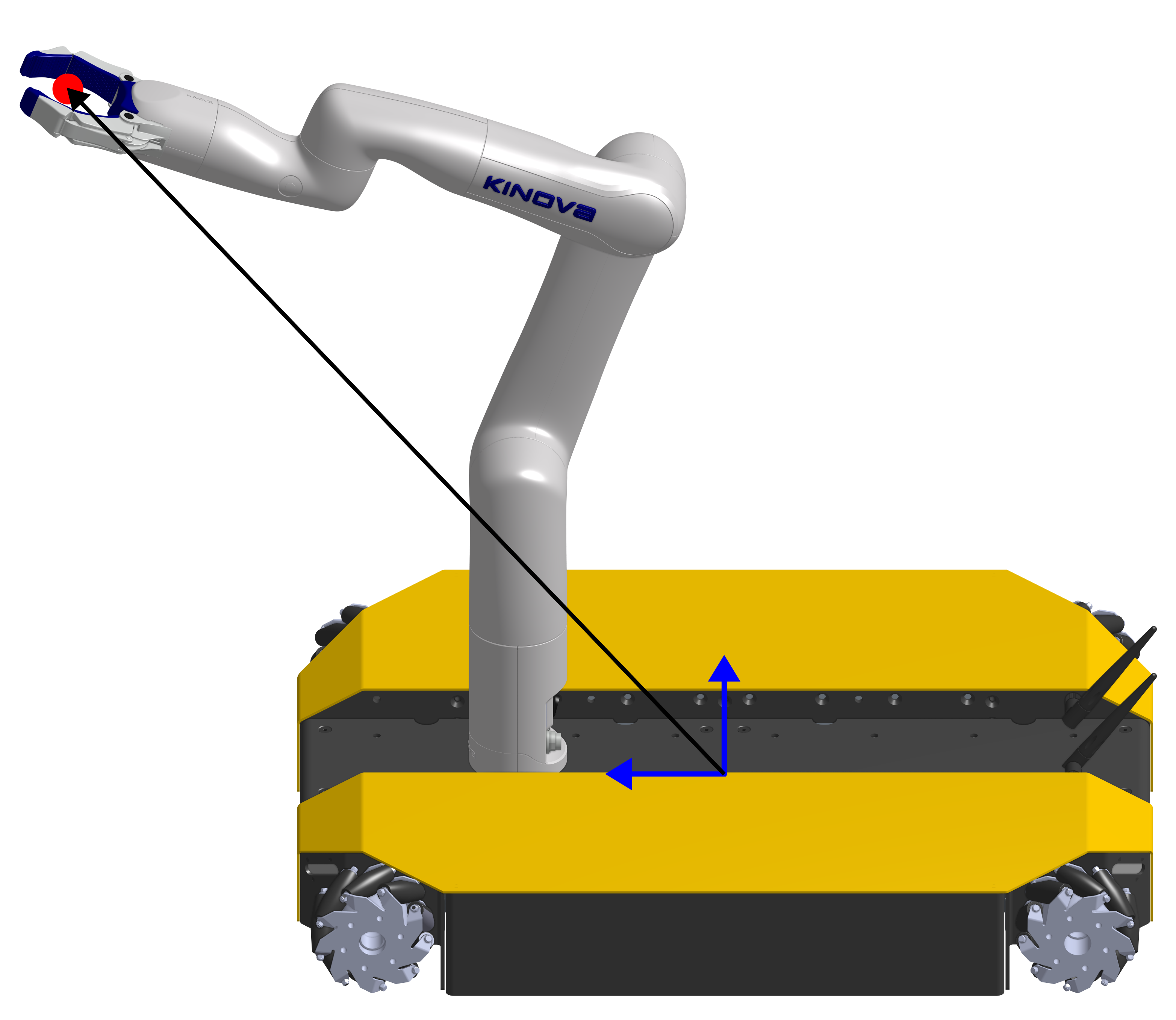}
        \caption{Static target in base frame.}
        \label{fig:guidance_mode}
    \end{subfigure}%
    \begin{subfigure}{.5\columnwidth}
        \centering
        \includegraphics[width=0.95\textwidth]{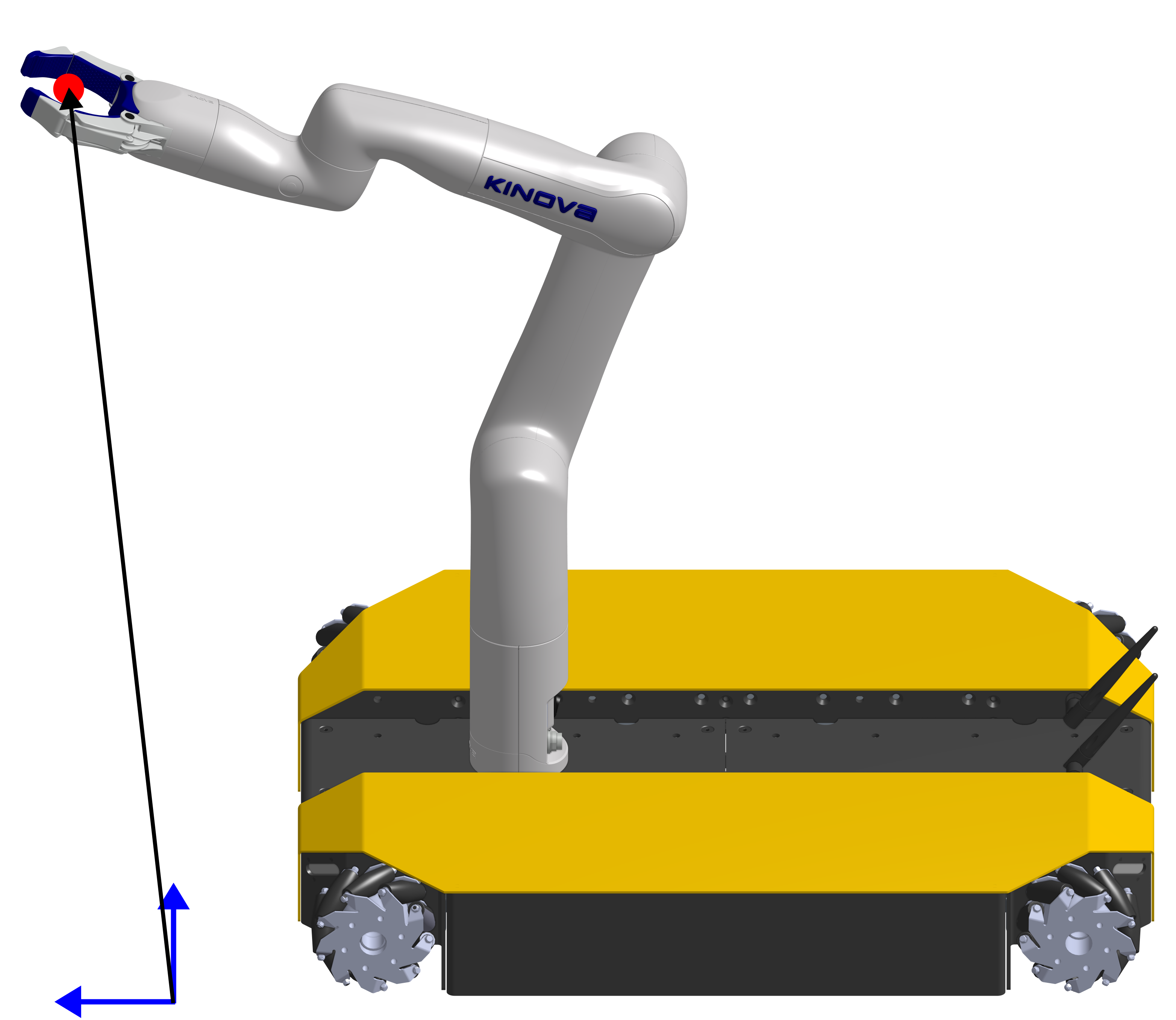}
        \caption{Target in world frame.}
        \label{fig:tracking_mode}
    \end{subfigure}%
    \caption[Guidance mode and tracking mode.]
    {Guidance mode (a) and tracking mode (b).}
    \label{fig:modes}
\end{figure}

\subsubsection{Guidance mode}
\label{subsection:guidance_mode}

In the guidance mode, the target is statically placed in the reference frame of the base, see \Cref{fig:guidance_mode}. By defining $\bm{p}_\mathrm{d}$ at the same location, the base will not move when the \gls{ee} reaches the target. When the user pushes the \gls{ee} away, the base will follow. Since the target is statically placed in the reference frame of the base, the target follows the movement of the base. When the \gls{ee} is released, it can return to the target, and the base stops. This setup allows the user to easily move the mobile manipulator by leading it through interaction with the arm.

\subsubsection{Tracking mode}
\label{subsection:tracking_mode}

In tracking mode, the target is placed in the reference frame of the world, see \Cref{fig:tracking_mode}. Therefore, the target does not follow the movement of the base, in contrast to the guidance mode. The user can push or pull the \gls{ee} away from the target, resulting in a movement of the base following the \gls{ee}. Since the target is defined in the reference frame of the world, the robot moves further away from the target as long as the user keeps pushing the robot away. When the robot is released, the \gls{ee} tries to get back to the target. The base will move, following the \gls{ee}, until the \gls{ee} is at the desired position $\bm{p}_\mathrm{d}$ relative to the base.

\section{EXPERIMENTAL SETUP}
\label{section:methodology}

All experiments are performed in the real world using the Kinova GEN3 Lite arm and Clearpath Dingo Omnidirectional
base. 

\begin{figure}[!b]
    \centering
    \includegraphics[width=0.9\columnwidth]{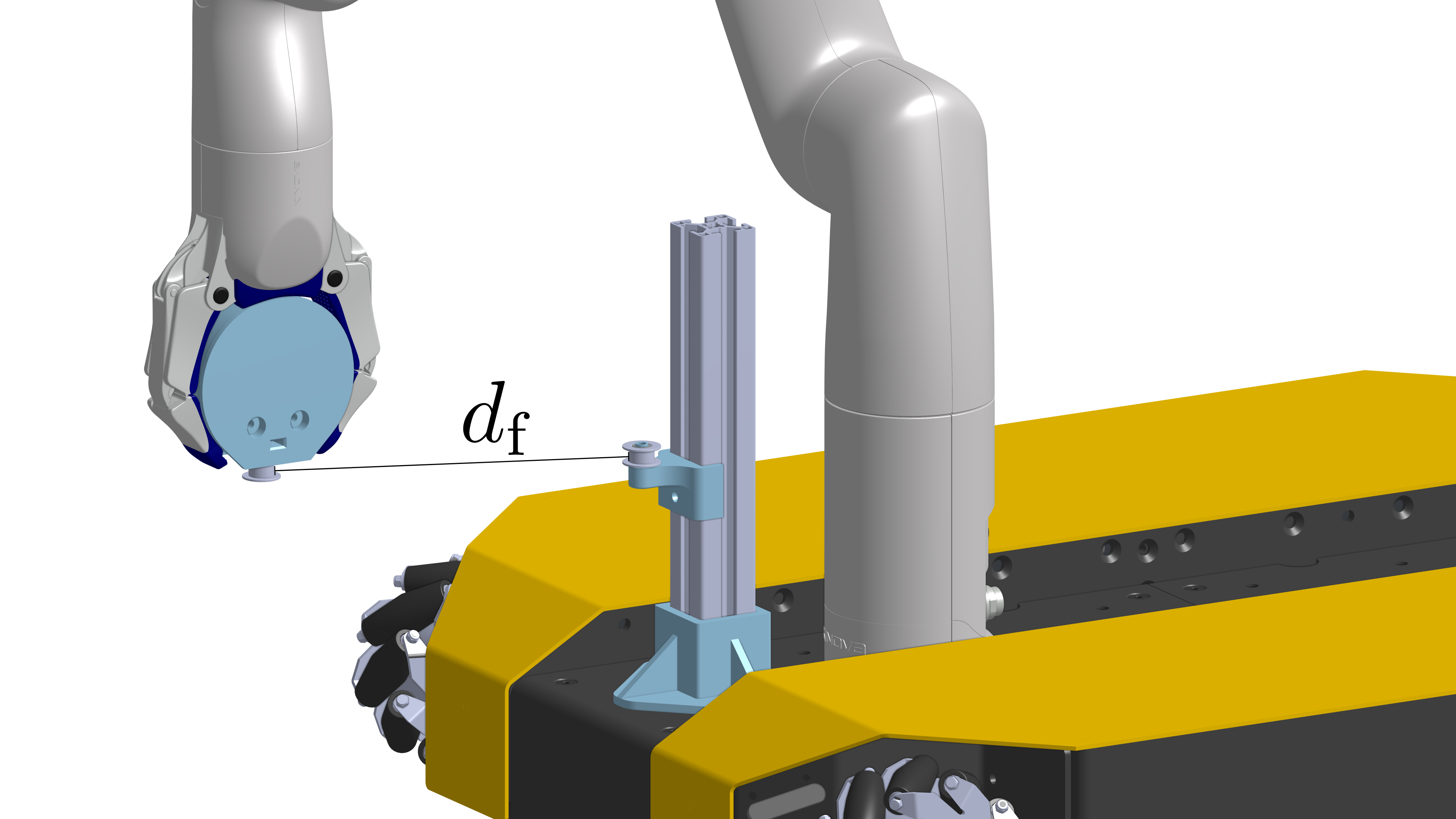}
    \caption[Setup of experiment 2.]
    {Setup of experiment 2. %
        A gripper handle and base connector are designed to limit the free distance $d_\mathrm{f}$ of the end-effector.}
    \label{fig:setup_experiment_1}
\end{figure}

\subsection{Experiment 1: consistency of the calibration}
\label{subsection:experiment_1}

This experiment validates the calibration method's consistency through $50$ repetitions. First, the phase shift between the real robot and its model is estimated, see \Cref{subsection:phase_shift}. Then, using the adjusted model, the current/torque ratio and friction loss are derived for all
joints affected by gravity, as described in \Cref{subsection:ratio} and \Cref{subsection:friction}. %
\subsection{Experiment 2: current-based impedance performance}
\label{subsection:experiment_2}

In this experiment, we evaluate the compliance of the adapted impedance controller by simulating interaction using springs with varying stiffness. To achieve this, a setup is arranged as illustrated in \Cref{fig:setup_experiment_1}. 
Between the handle in the \gls{ee}'s gripper and the extrusion profile, a rope containing a spring is attached, limiting the free-moving distance (where the spring is not extended) of the \gls{ee} to the length $d_\mathrm{f}$.

\begin{figure}[!b]
    \centering
    \includegraphics[width=0.9\columnwidth]{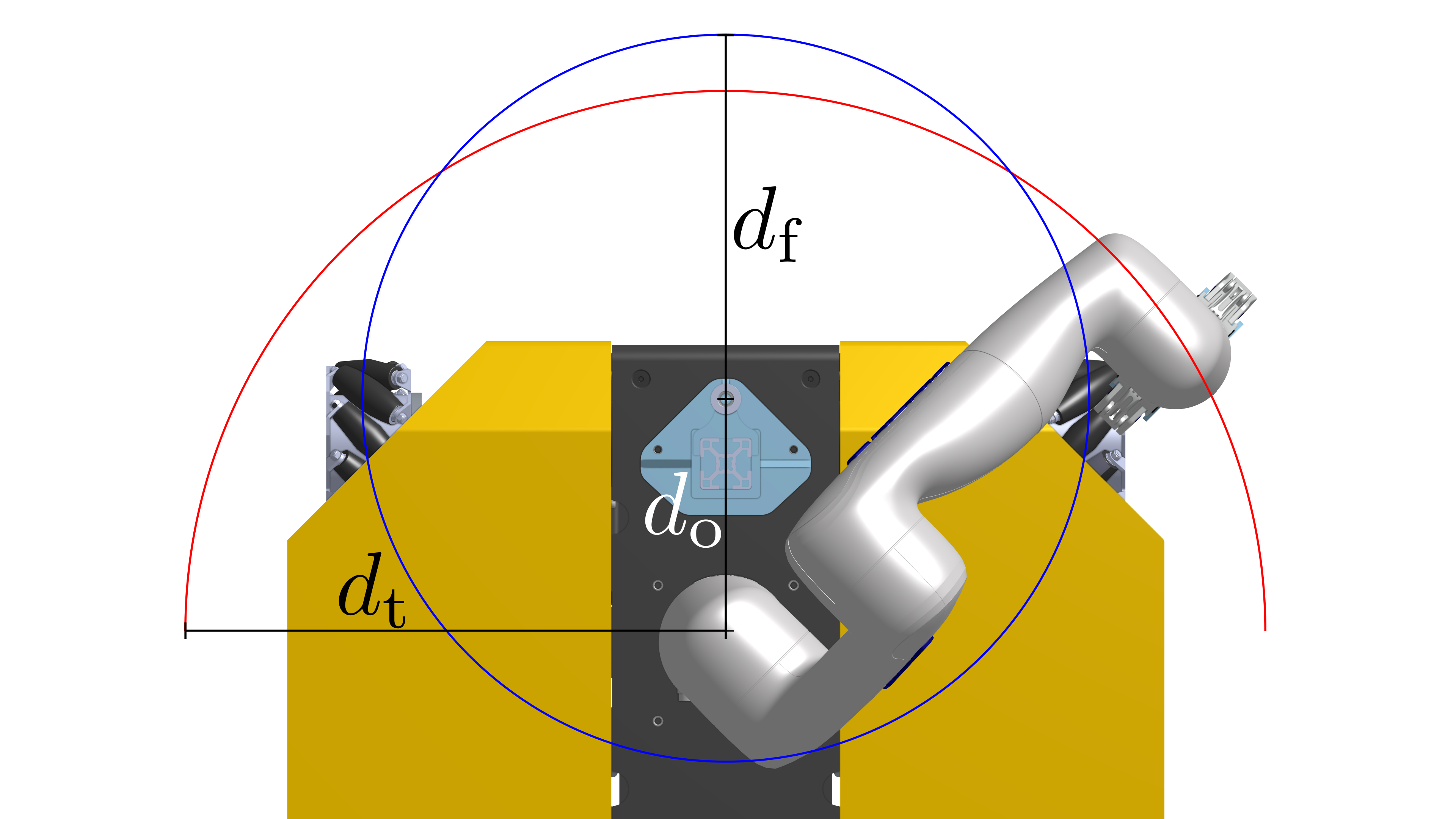}
    \caption[A spring limits the free-moving area of the robot.]
    {A spring limits the free-moving space of the robot. %
        The target trajectory (red) is half a circle with radius $d_\mathrm{t}$, %
        the free-moving space is within the blue circle with radius $d_\mathrm{f}$.}
    \label{fig:range_experiment_1}
\end{figure}

The task of the robot during this experiment is to track a target moving along a predefined
trajectory, as shown in \Cref{fig:range_experiment_1}. The target trajectory (red) is partly
within the range where the spring is unstretched (within the blue circle), and partly outside.
The part where the spring is not stretched simulates the situation of no physical interaction. In this range, the tracking performance of the compliant controller can be compared with
the default velocity controller. For the part where the spring stretches, the compliance of the arm
can be evaluated. The experiment is performed with a length of the unstretched rope $d_\mathrm{f} = 0.260
    \si{\meter}$, a target trajectory with radius $d_\mathrm{t} = 0.315 \si{\meter}$, a distance between the origins of the arm and the spring of $d_\mathrm{o} = 0.145
    \si{\meter}$ and a target velocity $v = 0.101
    \si{\meter\per\second}$. The target moves along the half-circle four times.
The experiment is repeated using springs of different stiffness, namely 0.01, 0.04, and 0.12 \si{\newton\per\milli\meter}.

Both the compliant and the velocity controllers have a
secondary nullspace task of containing a preferred configuration to avoid redundancy problems.  The designed impedance controller uses \Cref{eq:impedance_torque_ns_task} and \Cref{eq:impedance_controller}, with stiffness $\bm{K}_d = 40 \si{\newton\per\meter}$ and damping $\bm{D}_d = 3 \si{\newton\second\per\meter}$ and $\ddx_d = 0$, $\dx_d = 0$ for simplicity. \Cref{eq:controller_fc} is used for every joint to make the conversion from torques to currents.

\subsection{Experiment 3: mobile manipulator guidance and tracking}
\label{subsection:experiment_3}

This experiment combines the arm and the base, to validate the performance of the mobile
manipulator. We first test the guidance mode
defined in \Cref{subsection:guidance_mode} and then the tracking mode defined in
\Cref{subsection:tracking_mode}. The arm uses the same impedance controller as in
\Cref{subsection:experiment_2}, but  the error for the \gls{ee} is limited to a maximum of
0.1 \si{\meter} to avoid extreme forces when the target is far away from the \gls{ee}. Furthermore, a
different preferred pose is used as the null-space task to simplify the interaction with the robot. The base uses the controller described in
\Cref{subsection:base}.

Since the focus of this work is on the controller and since we aim to mitigate localization and perception errors, we utilize a virtual target and employ a motion capture system to monitor the robot's position.
\section{RESULTS}

This chapter presents the results of the experiments described in \Cref{section:methodology}.

\begin{figure}[!b]
    \centering
    \def\datafile{data/calibration/calibrate_model.csv}
    \includegraphics[width=\columnwidth, height=0.4\columnwidth]{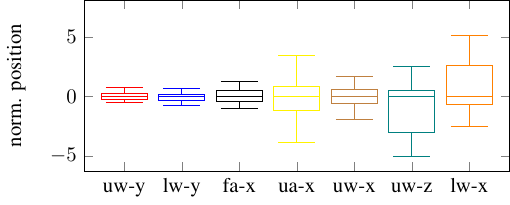}
    \includegraphics[width=\columnwidth, height=0.5\columnwidth]{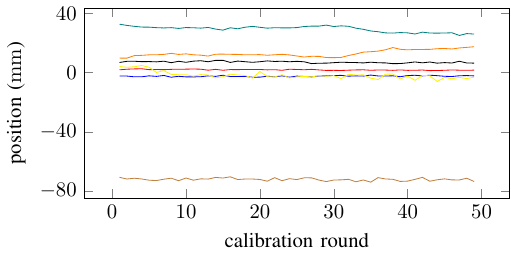}
    \caption{COM locations (link-direction) as boxplot normalized with the median and per calibration round.}
    \label{fig:results_com}
\end{figure}

\subsection{Experiment 1: consistency of the calibration}
\label{subsection:results_1}

In the first part of the experiment, the \gls{com} locations of the \gls{lw}, \gls{uw}, \gls{fa}, and \gls{ua} are estimated as outlined in \Cref{subsection:phase_shift}.
\Cref{fig:results_com} shows the results as a normalized boxplot and over the calibration
round. The length of each box being smaller than 4 mm showcases the consistency of the calibration
method. The most significant variances occur in the \gls{uw} z-location (cyan) and the \gls{lw} x-location (orange). Between the 30th and 35th calibration rounds, a synchronous shift becomes apparent in the two locations. The calibration of the former (cyan) impacts the latter (orange) due to the sequence of calibration.  This can explain the simultaneous change in the calibration outcome of
these two locations.

\begin{figure}[!b]
    \centering
    \def\datafile{data/calibration/calibrate_torque_and_friction.csv}
    \includegraphics[width=\columnwidth, height=0.4\columnwidth]{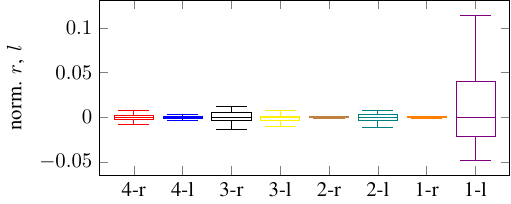} \\[0.1cm]
    \includegraphics[width=\columnwidth, height=0.5\columnwidth]{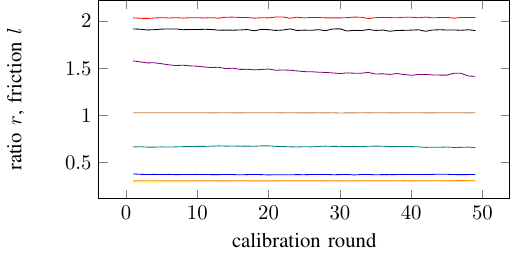}
    \caption{Joint (\#) ratio (r) and friction loss (l) as boxplot normalized with the median and per calibration round.}
    \label{fig:results_rf}
\end{figure}

In the second part of the experiment, the current/torque ratio~$r$ and the friction loss~$l$ are estimated for the joints experiencing gravity, namely joints 1 to 4. The robot model is adapted to include the average
\gls{com} locations derived above.  \Cref{fig:results_rf} shows the results as a normalized
boxplot and over time per calibration round. 
Except for the friction of joint~1~(purple), all parameters show a narrow interquartile range, demonstrating the
consistency of the calibration method. The plot per
calibration round shows that a continuous decrease in the friction of joint 1 is the reason for
the large variation. Joint~1 is the only joint containing a large actuator and needs to deliver the
highest torques since it is the first joint of the arm carrying all other joints and links. A
possible explanation for the decreasing friction might be the temperature changing over time during
the calibrations, but no temperature data is collected and therefore further research is
required to validate this theory.

Finally, the parameters of the two joints unaffected by gravity are computed as the average of those joints sharing the same actuator type. Note, that it is also possible to mount the arm in a different orientation to derive their parameters.

\subsection{Experiment 2: current-based impedance performance}
\label{subsection:results_2}

\begin{figure}[!t]
    \centering
    \includegraphics[width=1\columnwidth]{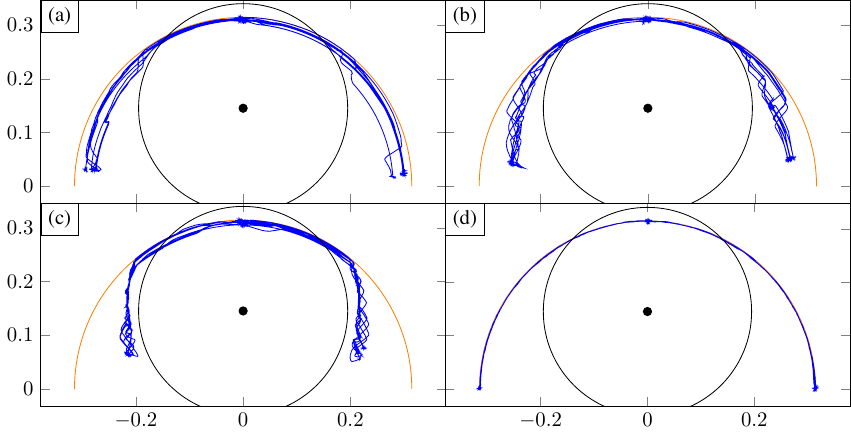}
    \caption[Target and actual trajectories for compliant and velocity control.]
    {Target (orange) and actual (blue) trajectories for compliant (a, b, c) and velocity (d) control. %
        Limited outside the black circles by different springs, a: 0.01, b: 0.04, c: 0.12, d: 0.12 \si{\newton\per\milli\meter}.}
    \label{fig:xy_tracking}
\end{figure}

The second experiment is designed to validate the performance of the current-based impedance
controller using the setup described in \Cref{subsection:experiment_2}. \Cref{fig:xy_tracking} shows the target trajectory
in orange and the actual trajectory of the \gls{ee} in blue, while the black circle indicates the
boundary at which the spring starts to pull the \gls{ee}. The compliant controller is used in
(a), (b), and (c) with springs of stiffnesses 0.01, 0.04, and 0.12 \si{\newton\per\milli\meter}, respectively. In (d), the velocity controller is used with the stiffest spring of 0.12
\si{\newton\per\milli\meter}.

The impedance controller is set up such that the stiffness of the \gls{ee} should be 0.04
\si{\newton\per\milli\meter}, similar to the stiffness of the spring in (b). It can be seen that the \gls{ee}
follows the orange target trajectory best in (a), since here the spring is weaker and thus more
compliant than the arm. In (b), the stiffness of the \gls{ee} should be equal to the stiffness
of the spring, which seems to be correct at first glance since the blue lines are about centered
between the orange line and the black circle, meaning that the \gls{ee} and the spring are equally strong. In (c), the stiffness of the \gls{ee} is smaller than the stiffness of
the spring, resulting in the \gls{ee} staying closer to the black circle, since it is
relatively more compliant than the spring.

Since the velocity controller is not compliant, regardless of the stiffness of the
spring, the result is equal to (d), where the stiffest spring
is used. The tracking performance can be evaluated inside the black circle, where no external forces are applied. The
velocity controller is able to track the target trajectory with a precision of less than 1 \si{\milli\meter}, while the compliant controller achieves a precision of about 5 \si{\milli\meter}.

\subsection{Experiment 3: mobile manipulator guidance and tracking}
\label{subsection:results_3}

The third experiment validates the performance of the complete mobile manipulator in the designed operational modes and can be seen in the provided video.

\Cref{fig:results_exp3} displays three frames of the robot being used in guidance mode. The left and center image, show the arm being pushed and pulled, respectively. Since the base follows the \gls{ee}, the base moves
until the arm can settle to its original configuration resulting in the base stopping. The right image
shows the compliant behavior of the mobile manipulator when someone bumps against the arm. 

Additionally, \Cref{fig:results_exp3} showcases three top-down frames illustrating the robot operating in tracking mode. The robot tracks a target (red) moving along the blue circle.  In the left image, the robot demonstrates compliance, allowing a human to move it while the target moves away. In the center image, the robot, released from human contact, repositions its \gls{ee} (green) back to the target. Due to the arm's extension toward the target, the \gls{ee} deviates from its default position relative to the base, prompting the base to adjust its position accordingly. Finally, in the right image, the \gls{ee} successfully reaches the target, enabling the robot to maintain tracking capability in the absence of interactions.
\section{CONCLUSION}
This paper presented a holistic approach to addressing compliance for mobile manipulators using off-the-shelf manipulators and wheeled mobile bases. Specifically, we introduced a calibration method that enables the application of impedance control on current-controlled manipulators. This method involves estimating the torque/actuator current ratio and friction loss and can account for model inaccuracies using only signals available from standard position control. Thus, no force/torque sensors are required. Furthermore, we presented two operational modes for interacting with the mobile manipulator that allow a human to guide the robot through interaction with the manipulator and extend the manipulator's workspace.
We showed the consistency of the calibration method in real-world experiments and we evaluated our impedance controller, which considers the identified torque/actuator current ratio and friction loss on a manipulator. 
We show the compliance of the impedance controller, which was able to track a target with a precision of about 5
 \si{\milli \meter} when not disrupted by external forces.
Additionally, we showcased the whole-body implementation on a mobile manipulator consisting of off-the-shelf components without relying on any force/torque sensors and showed that compliant control on a current-controlled robot without force/torque sensors is possible with good performance. 
 Future work can explore how the presented calibration method can be applied to a mobile base supporting current control.







\bibliography{references}

\begin{thebibliography}{10}
\providecommand{\url}[1]{#1}
\csname url@rmstyle\endcsname
\providecommand{\newblock}{\relax}
\providecommand{\bibinfo}[2]{#2}
\providecommand\BIBentrySTDinterwordspacing{\spaceskip=0pt\relax}
\providecommand\BIBentryALTinterwordstretchfactor{4}
\providecommand\BIBentryALTinterwordspacing{\spaceskip=\fontdimen2\font plus
\BIBentryALTinterwordstretchfactor\fontdimen3\font minus \fontdimen4\font\relax}
\providecommand\BIBforeignlanguage[2]{{%
\expandafter\ifx\csname l@#1\endcsname\relax
\typeout{** WARNING: IEEEtran.bst: No hyphenation pattern has been}%
\typeout{** loaded for the language `#1'. Using the pattern for}%
\typeout{** the default language instead.}%
\else
\language=\csname l@#1\endcsname
\fi
#2}}

\bibitem{zacharaki2020safety}
A.~Zacharaki, I.~Kostavelis, A.~Gasteratos, and I.~Dokas, ``Safety bounds in human robot interaction: A survey,'' \emph{Safety science}, vol. 127, 2020.

\bibitem{siciliano2008springer}
B.~Siciliano, O.~Khatib, and T.~Kr{\"o}ger, \emph{Springer handbook of robotics}.\hskip 1em plus 0.5em minus 0.4em\relax Springer, 2008, vol. 200.

\bibitem{bussmannWholeBodyImpedanceControl2018-05}
K.~Bussmann, A.~Dietrich, and C.~Ott, ``Whole-{{Body Impedance Control}} for a {{Planetary Rover}} with {{Robotic Arm}}: {{Theory}}, {{Control Design}}, and {{Experimental Validation}},'' in \emph{2018 {{IEEE International Conference}} on {{Robotics}} and {{Automation}} ({{ICRA}})}, 2018.

\bibitem{dietrich2016whole}
A.~Dietrich, K.~Bussmann, F.~Petit, P.~Kotyczka, C.~Ott, B.~Lohmann, and A.~Albu-Sch{\"a}ffer, ``Whole-body impedance control of wheeled mobile manipulators: Stability analysis and experiments on the humanoid robot rollin’justin,'' \emph{Autonomous Robots}, vol.~40, 2016.

\bibitem{kimMOCAMANMObileReconfigurable2020-05}
W.~Kim, P.~Balatti, E.~Lamon, and A.~Ajoudani, ``{{MOCA-MAN}}: {{A MObile}} and reconfigurable {{Collaborative Robot Assistant}} for conjoined {{huMAN-robot}} actions,'' in \emph{2020 {{IEEE International Conference}} on {{Robotics}} and {{Automation}} ({{ICRA}})}, 2020.

\bibitem{kimWholebodyControlNonholonomic2019-07}
S.~Kim, K.~Jang, S.~Park, Y.~Lee, S.~Y. Lee, and J.~Park, ``Whole-body {{Control}} of {{Non-holonomic Mobile Manipulator Based}} on {{Hierarchical Quadratic Programming}} and {{Continuous Task Transition}},'' in \emph{2019 {{IEEE}} 4th {{International Conference}} on {{Advanced Robotics}} and {{Mechatronics}} ({{ICARM}})}, 2019.

\bibitem{lamonIntelligentCollaborativeRobotic2020-05}
E.~Lamon, M.~Leonori, W.~Kim, and A.~Ajoudani, ``Towards an {{Intelligent Collaborative Robotic System}} for {{Mixed Case Palletizing}},'' in \emph{2020 {{IEEE International Conference}} on {{Robotics}} and {{Automation}} ({{ICRA}})}, 2020.

\bibitem{leboutetTactilebasedComplianceHierarchical2016-11}
Q.~Leboutet, E.~Dean-León, and G.~Cheng, ``Tactile-based compliance with hierarchical force propagation for omnidirectional mobile manipulators,'' in \emph{2016 {{IEEE-RAS}} 16th {{International Conference}} on {{Humanoid Robots}} ({{Humanoids}})}, 2016.

\bibitem{xing2021admittance}
H.~Xing, A.~Torabi, L.~Ding, H.~Gao, Z.~Deng, V.~K. Mushahwar, and M.~Tavakoli, ``An admittance-controlled wheeled mobile manipulator for mobility assistance: Human--robot interaction estimation and redundancy resolution for enhanced force exertion ability,'' \emph{Mechatronics}, 2021.

\bibitem{eom1998disturbance}
K.~S. Eom, I.~H. Suh, W.~K. Chung, and S.-R. Oh, ``Disturbance observer based force control of robot manipulator without force sensor,'' in \emph{Proceedings. 1998 IEEE International Conference on Robotics and Automation (Cat. No. 98CH36146)}, vol.~4.\hskip 1em plus 0.5em minus 0.4em\relax IEEE, 1998.

\bibitem{alcocer2003force}
A.~Alcocer, A.~Robertsson, A.~Valera, and R.~Johansson, ``Force estimation and control in robot manipulators,'' \emph{IFAC Proceedings Volumes}, vol.~36, no.~17, 2003.

\bibitem{jung2006robust}
J.~Jung, J.~Lee, and K.~Huh, ``Robust contact force estimation for robot manipulators in three-dimensional space,'' \emph{Proceedings of the Institution of Mechanical Engineers, Part C: Journal of Mechanical Engineering Science}, vol. 220, no.~9, 2006.

\bibitem{wahrburg2017motor}
A.~Wahrburg, J.~B{\"o}s, K.~D. Listmann, F.~Dai, B.~Matthias, and H.~Ding, ``Motor-current-based estimation of cartesian contact forces and torques for robotic manipulators and its application to force control,'' \emph{IEEE Transactions on Automation Science and Engineering}, vol.~15, no.~2, 2017.

\bibitem{liu2021sensorless}
S.~Liu, L.~Wang, and X.~V. Wang, ``Sensorless force estimation for industrial robots using disturbance observer and neural learning of friction approximation,'' \emph{Robotics and Computer-Integrated Manufacturing}, vol.~71, 2021.

\bibitem{wahrburg2018modeling}
A.~Wahrburg, S.~Klose, D.~Clever, T.~Groth, S.~Moberg, J.~Styrud, and H.~Ding, ``Modeling speed-, load-, and position-dependent friction effects in strain wave gears,'' in \emph{2018 IEEE International Conference on Robotics and Automation (ICRA)}.\hskip 1em plus 0.5em minus 0.4em\relax IEEE, 2018.

\bibitem{cao2022contact}
F.~Cao, P.~D. Docherty, and X.~Chen, ``Contact force estimation for serial manipulator based on weighted moving average with variable span and standard kalman filter with automatic tuning,'' \emph{The International Journal of Advanced Manufacturing Technology}, vol. 118, no.~9, pp. 3443--3456, 2022.

\bibitem{ott2008cartesian}
C.~Ott, \emph{Cartesian impedance control of redundant and flexible-joint robots}.\hskip 1em plus 0.5em minus 0.4em\relax Springer, 2008.

\bibitem{dietrich2021practical}
A.~Dietrich, X.~Wu, K.~Bussmann, M.~Harder, M.~Iskandar, J.~Englsberger, C.~Ott, and A.~Albu-Sch{\"a}ffer, ``Practical consequences of inertia shaping for interaction and tracking in robot control,'' \emph{Control Engineering Practice}, vol. 114, 2021.

\bibitem{dietrich2015overview}
A.~Dietrich, C.~Ott, and A.~Albu-Sch{\"a}ffer, ``An overview of null space projections for redundant, torque-controlled robots,'' \emph{The International Journal of Robotics Research}, vol.~34, no.~11, 2015.

\end{thebibliography}

\end{document}